\documentclass[10pt]{article}
\usepackage{arxiv}
\usepackage[utf8]{inputenc}  
\usepackage[T1]{fontenc}    
\usepackage{graphicx}
\usepackage[mathlines]{lineno}
\usepackage{cite}
\usepackage{amsmath,amssymb,amsfonts}
\usepackage{algorithmic}
\usepackage{textcomp}
\usepackage{xcolor}
\def\BibTeX{{\rm B\kern-.05em{\sc i\kern-.025em b}\kern-.08em
    T\kern-.1667em\lower.7ex\hbox{E}\kern-.125emX}}
\usepackage{amsmath, amsxtra, amsfonts, amssymb, amstext, mathtools}
\usepackage{booktabs}
\usepackage{makecell}
\usepackage{multicol}
\usepackage{multirow}
\usepackage{longtable}
\usepackage{array}
\usepackage{amsmath,mathtools}
\newcommand{\ignore}[1]{}

\DeclareMathOperator{\Var}{Var}

\usepackage{color}
\usepackage{xcolor}
\usepackage{colortbl}
\usepackage{url}
\usepackage{hyperref}

\begin{document}

\title{Improving Confidence in Evolutionary Mine Scheduling via Uncertainty Discounting}

\author{
Michael Stimson\\
Maptek Pty. Ltd\\
Adelaide, Australia\\
 \And
William Reid\\
Maptek Pty. Ltd\\
Adelaide, Australia\\ 
\And
Aneta Neumann\\
Optimisation and Logistics\\
The University of Adelaide\\
Adelaide, Australia\\
\And
Simon Ratcliffe\\
Maptek Pty. Ltd\\
Adelaide, Australia\\
\And
Frank Neumann\\
Optimisation and Logistics\\
The University of Adelaide\\
Adelaide, Australia\\
}

\date{}
\maketitle

\begin{abstract}
Mine planning is a complex task that involves many uncertainties. During early stage feasibility, available mineral resources can only be estimated based on limited sampling of ore grades from sparse drilling, leading to large uncertainty in under-sampled parts of the deposit. Planning the extraction schedule of ore over the life of a mine is crucial for its economic viability. We introduce a new approach for determining an ``optimal schedule under uncertainty" that provides probabilistic bounds on the profits obtained in each period. This treatment of uncertainty within an economic framework reduces previously difficult-to-use models of variability into actionable insights. The new method discounts profits based on uncertainty within an evolutionary algorithm, sacrificing economic optimality of a single geological model for improving the downside risk over an ensemble of equally likely models. We provide experimental studies using Maptek's mine planning software Evolution. Our results show that our new approach is successful for effectively making use of uncertainty information in the mine planning process.
\end{abstract}

\keywords{Evolutionary algorithms, open pit mine optimization, open pit mine production scheduling, uncertainty, mine planning}

\section{Introduction}
Long-term planning and production scheduling are among the most critical tasks in the area of mining. The goal is to extract valuable ore from an orebody in a sequence that takes into account many mining and precedence constraints in a way that is economically efficient \cite{bienstock2010solving}. This is an important real-world optimisation problem that has been studied in the literature over many years.
This includes mixed integer programming approaches based on block scheduling~\cite{DBLP:journals/coap/MunozEGMQL18,DBLP:journals/ior/LetelierEGMM20}. Each block in a block model (a discretised spatial approximation of the mineral deposit) has a given estimated value based on the metal grade and the excavation cost. Other heuristic techniques include dealing with specific characteristics such as uncertainties of the problem~\cite{DBLP:journals/asc/GoodfellowD16,DBLP:journals/heuristics/MontielD17,DBLP:journals/cor/LamghariD20}. Different software products that offer a variety of approaches for mine planning and extraction sequences are available~\cite{maptek,minemax}.

Evolutionary computation techniques have successfully been applied in the area of mining, 
in particular to large scale optimisation problems such as the cost
efficient extraction of ore \cite{DBLP:journals/jour/Ibrahimov2014,DBLP:conf/cec/OsadaWBM13}, the ore processing and blending problem \cite{chakraborty2012multi,DBLP:conf/gecco/SchellenbergLM16,DBLP:conf/gecco/XieN022,xie2021heuristic,DBLP:conf/cec/Xie0N21}, and the large-scale open pit mine scheduling problem \cite{myburgh2010evolutionary,elsayed2020evolutionary}. Particle swarm algorithms were utilised to solve the capacity constrained open pit mining problem \cite{ferland2007application} and the transportation and layout problem of locating a crushing station in an open-pit mine \cite{gu2021layout}.

\subsection{Uncertainty in Mine Planing}
Uncertainty is a feature common to many financial investments. For example, when investing in shares there is uncertainty around the future performance of the stock. This creates considerable risk as well as the potential for big reward. Along the same lines, uncertainty around the true composition of mineral deposits comes with considerable financial risk when investing in a mining operation.

The only way to know the true value of a mineral deposit is to mine it entirely. At this point, the money to dig and process the material has already been spent and the downside risk or upside reward has been realised. To reduce this potential for large expenditure in the face of unknown returns, deposits are mapped through a programme of drilling. Samples of the drilled material at known depths down holes carefully planned in known locations are chemically assayed. This provides estimates of the composition of the ore within the orebody.

With a large collection of assayed holes distributed spatially throughout the deposit of interest, estimations can be made about what type and grades of material might exist throughout the entirety of the rock volume between the holes. This process is typically used to produce a block model. One such estimation process is called Kriging \cite{cressie1990origins,kleijnen2009kriging}, a method of interpolation based on Gaussian processes governed by prior co-variances over a set of spatially distributed sample measurements. Kriging is deterministic and so produces the same result for the same input data. It fails to capture the uncertainty present in the estimations.

Increasingly, methods are being adopted that better capture the uncertainty by using stochastic estimation algorithms and running them many times to produce an \emph{ensemble} of equally probable block models. One broad category of such modelling techniques is conditional simulation \cite{hoshiya1995kriging,leuangthong2003stepwise}. Another newer approach is emerging from machine learning that models training data slightly differently each time from a randomised starting point. No matter the method, the thing that is common to them is that a spatially varying measure of uncertainty is captured across the ensemble of block models produced.

Traditionally, a single resource block model is used in an economic feasibility study to determine if a deposit is worth mining. Feasibility follows from economic assumptions of input costs, expected selling prices and the time value of money modelled as a discount rate. These assumptions guide the mine planning process which determines which parts of the resource to mine, which of those to process as ore versus waste (cut-off grade policy) and the sequence and timing of mining and processing them. This is a well studied optimisation problem that seeks to maximise the NPV of a mineral resource \cite{DBLP:journals/coap/MunozEGMQL18,DBLP:journals/ior/LetelierEGMM20}. Very large capital investments hinge on the recommendations of such studies and they are often presented to investors to secure loans. If the real NPV of the deposit is well below what a study based on a given resource block model claims, by the time you learn this it is too late. The money is spent, the non-economic ore is mined and you've still got a large loan to repay!

\subsection{Our contribution}
Techniques that can effectively utilise an ensemble of block models rather than a single block model for determining economic feasibility are required to improve the traditional approach. Previous work in this area \cite{DBLP:conf/gecco/Reid0RN21} provided a method for visualising the uncertainty on per mining period graphs of NPV and profit. This provided a quantitative assessment of how geological uncertainty manifested as economic uncertainty. Attempts to use this information to change the mining sequence manually to reduce downside risk (improve the worse solutions up towards the mean) were only modestly effective. We now present a technique that uses the uncertainty information directly in optimisation of NPV. This changes the schedule and cut-off grade policy used based on observed uncertainty and spatial correlations. It optimises simultaneously across all block models in the ensemble. This has the effect of improving the worse-case solutions at the expense of the mean solution. The result is a more conservative estimate of NPV which can be made with quantified greater confidence. It can be readily compared with traditional methods.

In Section~\ref{sec:key} and ~\ref{sec:approach}, we define key concepts and describe the intended approach that encourages blending of certain and uncertain material together. In Section~\ref{sec:theo}, we describe the theoretical motivation for the discounting process. We provide the problem definition used in Maptek's Evolution software, and introduce uncertainty qualification based on discounted profit in Section~\ref{sec:Mine Planing}. We present and discuss our experimental results of our new approach for reducing uncertainty in Section~\ref{sec:test_model} and a more complex real world model in Section~\ref{sec:complex_model}. Finally, we finish with some concluding remarks.
\section{Key concepts}
\label{sec:key}
A resource model is stored as a set of attributes within a regular block model. Typical attributes are rock lithological domain (a codified representation of the material's geological provenance), grade (the concentration of an ore, an element of economic interest expressed as a percentage or part per million within that rock) and rock density to facilitate calculation of both rock and ore tonnages. An ensemble block model contains multiple realisations (sets of attributes) of grade and possibly lithological domain and density. Each ensemble member is a single realisation of the resource model. Each contain the same number of blocks and attributes but individual blocks will differ in grade and possibly lithological domain and density, due to the potential for variation in the way each realisation is created.

A resource model is typically grouped into blocks that will be mined and those that will be left. The set of blocks that will be mined form a pit or pits and are commonly \emph{staged}. This term is used to denote the grouping and ordering of large subsets of blocks within the pits involved into separate mining campaigns \cite{lerchs-grossmann,caccetta2007application,whittle1991open}. Stages are often designed to be followed in sequence, i.e stage 1 is mined out before any mining can start in stage 2. This is a common practice however it does not need to be explicitly adhered to, provided it does not violate any geometrical constraints. A stage is \emph{open} if it is available for mining.

A bench is the set of blocks at a given height or level within the block model. A stage-bench combination is all the blocks at a given level within a given stage. A stage-bench-domain combination is all the blocks at a given level within a given stage belonging to the same lithological domain. A stage-bench-domain-bin combination is all the blocks at a given level within a given stage belonging to the same lithological domain with a material grade between a given upper and lower bound.

Maptek Evolution Strategy is commercial software for optimising the extraction sequence of material to maximise NPV. It does not operate on individual blocks but rather on stage-bench combinations of blocks and then on domain-bin combinations within them. The material movements required and the economic returns won are aggregated into (typically) yearly periods from a sequence of such combinations giving a means of comparison between different sequences and thus a function to optimise. Reid et al.~\cite{DBLP:conf/gecco/Reid0RN21} discuss this approach in further detail.

\subsection{Illustrating key concepts in a test model}
\begin{figure*}[t!]
\centering
\includegraphics[width=1\textwidth]{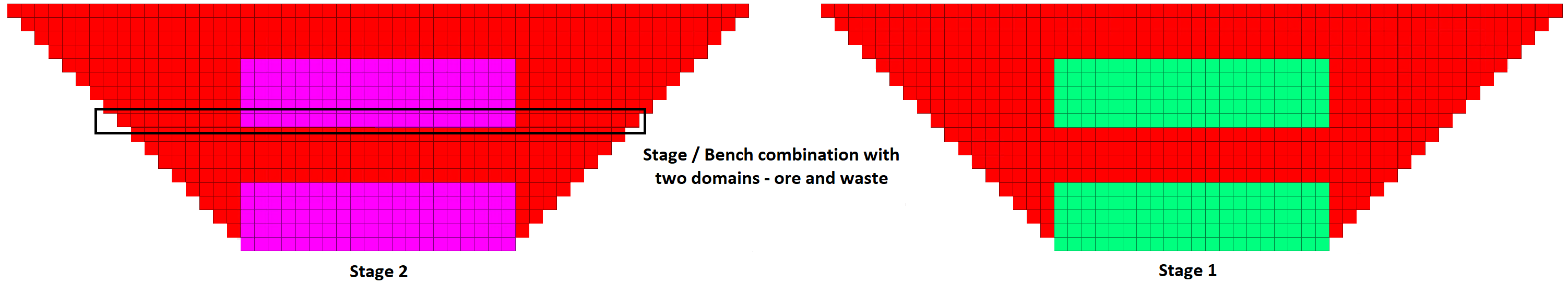}
\caption{(Test model) A cross section in the Y plane, showing copper ore in two separate pits, each containing two separate box shapes.}
\label{fig:quality}
\end{figure*}

A test model shown in Figure~\ref{fig:quality} helps illustrate these concepts. It consists of two identically shaped pits each containing two uniform `boxes' of ore comprising $2000$ blocks. The boxes in each pit are separated by a layer of waste material that must be mined to expose the lower box. Additional waste blocks are also present to honour a $45$ degree pit slope angle creating a total of $18,520$ waste blocks to go with the $8,000$ ore blocks. All blocks in the waste domain are assigned a grade of zero. The pits are stages by simply denoting the East (right) pit as stage 1 and the West (left) pit as stage 2.

The grade values assigned to the ore blocks vary across 10 ensemble members. Those in the East pit are given a constant grade for each ensemble member with values uniformly distributed in the range 
$[0.46\%, 0.55\%]$ and are monotonically decreasing with each member of the ensemble. Those in the West pit likewise except that the grades are uniformly distributed in the range $[0.41\%, 0.5\%]$ monotonically \emph{increasing} through the ensemble. The result is an ensemble model that has a grade distribution that is positively correlated between benches in both pits and negatively correlated between each pit across the ensemble, with the East pit having slightly more valuable ore.
\section{Intended approach}
\label{sec:approach}
Reid et al.~\cite{DBLP:conf/gecco/Reid0RN21} discussed approaches to reduce uncertainty in a mining schedule given an ensemble of equally probably reserve models. The paper looked at different manual staging approaches in detail and although it showed that the position of stage boundaries and the order in which stages were mined could result in changes to the uncertainty of NPV and the uncertainty of per period profits, no clear pathway was identified to use these insights to consistently reduce the downside risk (the performance of the worst performing members of the ensemble) due to uncertainty. 

Uncertainty in an ensemble reserve model is defined as the grade variability present in individual blocks and subsequent aggregations of these blocks into blobs of blocks. The mining of blocks with high variability, no matter where they are located within the ultimate pit, has to be accepted at some point in the mining schedule. Our aim is to place them in the mining schedule in a way that is in some sense optimal. A naive optimal solution to mitigate the effect of uncertainty on NPV would be to mine material with lowest uncertainty (those blocks exhibiting lowest variability across all members of the ensemble) at the start of the schedule graduating through to material with the highest uncertainty (those block exhibiting the highest variability across all members of the ensemble) at the tail end of a schedule. In reality, this is almost always impractical due to geometric constraints and would also potentially mine a lot of high grade ore late in the schedule having a large downside effect on the mean NPV of the ensemble.

A better approach may be to try to blend uncertainty away. This can be achieved in two ways: either blend material of high variability with material of low variability or blend materials of high variability together provided the variability is not correlated. The latter method exploits the property that aggregating a number of parcels of material together with uncorrelated variability will result in a new larger parcel whose uncertainty is lower, trending to the volume weighted mean of the input parcel uncertainties.

This blending of material can help to reduce some of the downside risk associated by ensuring some certain cash flow is going to occur during mining. This approach is difficult to do manually. Since stages typically contain material for multiple periods, there is no guarantee uncertain material will be blended with certain material in the right quantities to guarantee good economic performance in each period.

This led us to focusing on the scheduling algorithm utilised by Maptek Evolution Strategy. In an attempt to blend certain and uncertain material together, the evolutionary algorithm's fitness function was targeted to downweight scheduling sets of stage-bench-domain-bin parcels based on the uncertainty of each parcel and the correlation of this uncertainty between these parcels. Our hypothesis is that this approach would result in an overall lower mean NPV across the ensemble, but also a reduction in the per-period profit variance, thus leading to a more economically conservative mining sequence but one with which mining engineers could be confident was more resilient to downside risks arising from uncertainty.

\section{Theoretical Motivation}
\label{sec:theo}
We now provide a theoretical motivation for the discounting process that we will later use for mine planning. To do this, we guarantee profits based on the uncertainty measured in terms of the variance of a solution.

Let $p_i$, $1 \leq i \leq n$, be random profits of the set of $n$ parcels of material from which we can choose. The search space is $\{0,1\}^n$ and parcel $i$ is chosen by a search point $x$ iff $x_i=1$ holds.

We follow the approach taken in~\cite{DBLP:conf/ppsn/NeumannXN22} for the knapsack problem with stochastic profits. 
Our goal is to compute a solution for which we can guarantee a profit of at least $P$ with probability at least $\alpha$ where we aim to maximize $P$. Hence, we want to find a solution $x \in \{0,1\}^n$ with
\begin{eqnarray}
\label{eq:maxP}
   & \max P \text{~~~~ s.t } ~~~Pr(p(x) \geq P) \geq \alpha.
\end{eqnarray}

Throughout this paper, we assume $\alpha \in[1/2, 1[$.

We now discuss fitness functions motivated by the investigations carried out in~\cite{DBLP:conf/ppsn/NeumannXN22}.
and assume different types of distributions for the profits.  We denote by $\mu_i$ the expected profit, by $\sigma_i^2$, the variance of the profit of parcel $i$, $1 \leq i \leq n$. 
Let $E(x) = \sum_{i=1}^n \mu_i x_i$ be the expected profit and $Var[x] = \sum_{i=1}^n \sigma_i^2 x_i$ the variance of the profit of a solution.

If we assume that the $p_i$ are independent and chosen according to a Normal distribution $N(\mu_i, \sigma_i^2)$, then according to~\cite{DBLP:conf/ppsn/NeumannXN22}
the maximum of $P$ is equal to the maximum of 
\begin{equation}
\label{eqn:profit-discount-normal}
    f_{Norm}(x) =  E(x) - F_{\alpha} \cdot \sqrt{Var[x]}.
\end{equation}
Here $F_{\alpha}$ is the $\alpha$-fractional point of the standard Normal distribution.
Hence, solving problem (\ref{eq:maxP}) is equivalent to maximizing the fitness function~(\ref{eqn:profit-discount-normal}).

Usually, the profits among parcels are not independent and there is some correlation especially if they are spatially proximate.
We denote by $Cov(i,j)$, the covariance of parcel $i$ and $j$, $1 \leq i < j \leq n$.
Note that we have $Cov(i,j)>0$ if $i$ and $j$ are positively correlated, $Cov(i,j)<0$ iff $i$ and $j$ are negatively correlated, and $Cov(i,j) = 0$ if there is no correlated between parcels $i$ and $j$.

For the variance of $x$ we have 
$$Var[x] = \sum_{i=1}^n \sigma_i^2 x_i + 2 \sum_{1\leq i<j \leq n} Cov(i,j)x_ix_j.
$$

Assuming that we can determine or estimate the expected value and the variance of a solution based on the ensemble data, we can use the one-sided Chebyshev inequality to quantify the impact of uncertainty.

Following the calculations using Chebyshev's inequality and taking into account the expected profit and variance as done in~\cite{DBLP:conf/ppsn/NeumannXN22},
we may use the fitness function

\begin{equation}
\label{eq:cheb}
f_{Cheb}(x) = E[x] -  C_{\alpha} \cdot \sqrt{\Var[x]}.
\end{equation}
where $C_{\alpha}=\sqrt{\alpha/(1-\alpha)}$ based on Chebyshev's  inequality to discount the profit of a solution.
It can be observed that both Equation~\ref{eqn:profit-discount-normal} and Equation~\ref{eq:cheb} discount the expected profit of a solution $x$ by a constant factor (dependent on $\alpha$) times the standard deviation of the profit of solution $x$.
We follow this approach when dealing with uncertainties in mine planning and discount expected profits by a weightening of its variance. 

So, we consider fitness functions of the form
$$f(x) = E[x] -  K_{\alpha} \cdot \sqrt{\Var[x]}$$
where $K_{\alpha}$ is a constant based on the desired confidence level $\alpha$. Note, that increasing the value of $K_{\alpha}$ leads to a more conservative approach where solutions with a smaller standard deviation are increasingly preferenced over those with a higher expected value.

\section{Mine Planing using uncertainty adjusted profits}
\label{sec:Mine Planing}
The previous theoretical investigations motivate how we downgrade profit based on the uncertainty of a solution. In the following, we present the downgrading process for the mine optimisation problem and discuss its implementation details.

\subsection{The Evolution Strategy fitness function}
Evolution Strategy is Maptek’s long term mine scheduling package. It is a single objective genetic algorithm which attempts to maximize the NPV of a user configured mine site. The fitness of a solution in the population is a calculation of its NPV. The details of this calculation are as follows.
\begin{itemize}
\item Each parcel $b$ of material is attributed with a mass $m(b)$ and grade $g(b)$ by aggregating its component blocks from a single block model.
\item Time periods $T$ allow for changes in user configuration as well as allowing profit to be discounted over time (NPV)
\item Whenever a parcel is mined, a mining cost $n(b, t)$ is incurred, where $t$ is the period in which mining occurs.
\item When a parcel is sent to waste, a rehabilitation cost $h(b)$ is incurred.

\item When a parcel is processed, a processing cost is $q(b, t)$ incurred, where $t$ is the period in which processing occurs.

\item When a parcel is processed, only a portion $r(b)$ of the available metal is recovered because the recovery of the mill is not perfect - some metal is lost to the mill waste stream.

\item When an amount of metal is sold, a selling cost $c(b)$ is incurred which is a function of both $m(b)$ and $g(b)$.

\item When an amount of metal is sold, it is sold for a price $i(b, t)$, where $t$ is the period in which the selling occurs.

\end{itemize}

The saleable metal produced from a parcel sent to the mill is
$
l(b) = m(b) \cdot g(b) \cdot r(b).
$
The value of processing a parcel is
$
v(b,t) = l(b) \cdot (i(b, t) – c(b, t))  -  m(b) \cdot q(b, t).
$

Finally, the total profit of the parcel $b$ that is processed is given as 
$
p(b, t) = v(b, t) – (m(b)  \cdot n(b, t)).
$
The profit of a parcel $b$ sent to waste is given as
$
p(b, t) =  -m(b)  \cdot (n(b, t) + h(b)).
$
Note that the profit is negative if the parcel is sent to waste.

The fitness function given a sequence of all parcels in the reserve is the sum of all per-parcel costs and profits discounted by the periods in which these profits and costs were incurred, calculated as 
$$
f(B) = \sum_{(b,t) \in B} \frac{p(b,t)}{ ((1 + D)^{t})}
$$
where D is the yearly discount rate. This quantity is the NPV (net present value) of the reserve.

\subsection{Modifications to the Evolution Strategy fitness function}

The first part of our approach is to change the fitness function of the genetic algorithm. We replace the profits in our standard NPV calculation with profits downrated by an uncertainty risk term
$
p^*(b,t) =  p(b, t) – UR(X,t).
$

For the uncertainty risk $UR(X,t)$, we need to define some additional properties of the material mined in a given period. Let $X = \{b_1, \ldots, b_n\}$ be the set of $|{X}|$ parcels mined in a given period that are delivered to the mill. Let $E = \{e_1, \ldots, e_m\}$ be an ensemble of $|{E}|$ equally probably estimates of the grade distribution across all parcels. For each member of the $E$ each parcel in $X$ will be stamped with a grade, denoted $g(b, e)$. This can be used to calculate a profit value for that parcel $p(b, t, e)$ in time period $t$ using the same formulation as above on an ensemble memberwise basis:
$$
l(b, e) = m(b) \cdot g(b, e) \cdot r(b)
$$
$$
v(b, t, e) = l(b, e) \cdot (i(b, t) – c(b, t))  -  (m(b) \cdot q(b, t) )
$$
$$
p(b, t, e) = v(b, t, e) – (m(b)  \cdot n(b, t)).
$$

The mean and variance in profit across the ensemble for parcel $b$ in period $t$ are given by
$
\mu(b,t) = \frac{1}{|{E}|} \cdot \sum_{e \in E} p(b,t,e)
$
and
$
\sigma^2(b,t) =   \frac{1}{|{E}|} \cdot \sum_{e \in E}  (p (b,t,e) - \mu(b,t))^2.
$

Our uncertainty risk consists of two terms, one for the ensemble variance in profit for each parcel in X and one for ensemble covariance in profits across all parcels in X. The variance term
is
$
Var(X,t) = \sum_{b \in X}  \sigma^2(b,t).
$
and the covariance term 
is
$
Cov(X,t) = \sum_{b, b' \in X, b\not=b'} Cov(b, b', t), 
$
where
$$
Cov(b, b', t) = \frac{1}{|{E}|} \cdot \sum_{e \in E} (p(b,t,e) - \mu(b,t)) \cdot (p(b',t,e) - \mu(b',t)).
$$

Although negatively correlated grades are not expected from an in situ block model in the mining context, we clamp this covariance to have a minimum of zero so that negatively correlated material cannot be “uprated”. We have
$
Cov^*(X, t) = \max\{0, Cov(X, t)\}.
$

Combining the variance and covariance terms in a manner consistent with Chebyshev's inequality, we have
$
SV(X, t)  = \sqrt{Var(X, t) + Cov^*(X, t)}.
$
We refer to this value as the Standard Variance measure for profit for a given set of parcels $X$ milled in period $t$. By summing over all periods in a given schedule we obtain the total Standard Variance of a schedule, a measure of its profit uncertainty independent of NPV, duration or other factors. When comparing schedules, a lower total Standard Variance implies a more certain schedule. We have
$
SV(B)  =  \sum_{t \in T} SV(X, t).
$

A weighted term incorporating the Standard Variance is used to calculate an uncertainty risk $UR(X,t)$ for each parcel in $X$ due to uncertainty as:
$$
UR(X, t)  =   \frac{F_{\alpha}}{|X|} \cdot SV(X, t)=  K_{\alpha} \cdot SV(X, t)
$$
where $K_{\alpha} =  \frac{F_{\alpha}}{|X|}$. In our experiments were are using values for $\alpha$ of $60\%$, $90\%$, and $99\%$ which implies values of $F_{\alpha}$ of approximately $0.25$,  $1.28$, and $2.32$, respectively. 

This allows us to downrate the profit of a parcel using
$
p^*(b,t) =  p(b, t) – UR(X,t)
$
and adjust the overall NPV calculation to incorporate the uncertainty downrated profits to give the final revised fitness function as
$$
f(B) = \sum_{(b,t) \in B} \frac{p^*(b,t)}{ ((1 + D)^{t})}.
$$

\subsection{Overview of the Evolution Strategy GA}

For this paper, we assume that a single element is being targeted by the mining operation. A configured mine site contains a block model of the mineable material as well as the potential destinations for mined material. The genome of an individual solution is the combination of a set of cutoff grades for the element involved for each destination as well as a simplified material extraction sequence (blocks are aggregated into user defined stage-bench combinations and valid schedules comprise an ordering of such combinations that do not violate geometrical constraints). Strategy evaluates the fitness of a genome by first decoding it into the attributed parcels of material that end up at each destination in each year based on the domain-bin parcels found within each stage-bench.

The initial extraction sequences of material are generated using a biased random selection from among the available material to be mined. The more valuable the material, the more likely it is to be selected. The initial cutoff grades (bin boundaries for each domain) of the process are generated normally distributed around the breakeven cutoff of the mill (the theoretical grade of material at which the process makes money). The cutoff grades are mutated through a similar method, randomly moved with probability determined by a normal distribution centered at the current value.

In general, an extraction sequence is optimal if it succeeds at filling the mill in each year of the mine life without mining any additional ore in that year.

\subsection{Modifications to the Evolution Strategy GA}
The second part of our approach is to improve the sequence spawner to create sufficient genetic diversity in the population under the new fitness evaluation function. Because the fitness of individuals in the population now incorporates uncertainty, greater diversity in the sequences of population members is needed to explore the solution space for optimality. A justification for this observation is that mining sequences that combine widely separate materials from various pits into parcels are now potentially rewarded significantly more than was the case in the simple NPV fitness evaluator so a spawner that introduces such sequences is needed. The new behaviour:
\begin{itemize}
\item Maintains a set W of the last w benches mined
\item Maintains a set of the available benches to mine R
\item For each possible bench r in R, determine the uncertainty discounted profit of the set of benches W + r
\item Makes a bias random selection of r in R weighted towards the most profitable set W + r
\item Adds r to the extraction sequence and to W, remove the first bench in W to maintain its size. 
\end{itemize}

This new spawner will create some extraction sequences that heavily incentivise blending uncorrelated material together. If used in combination with the original spawner, some sequences in the population will “chase high NPV” and some sequences will “blend uncorrelated material”. As the population evolves, the traits of these sequences will be combined together to generate the theoretically optimal extraction sequence that strikes the optimal balance between chasing high NPV and uncorrelated blending to mitigate uncertainty.

\begin{table*}[!t]
\small
\centering
\caption{\label{tab:1}
Maximum (\text{max}), minimum (\text{min}), mean (\text{mean}), and standard deviation (\text{std}) in terms of expected value for NPV $(\$)$ with a $8\%$ discount rate for Strategy and uncertainty discounted Strategy with $8\%$ discount rate, and a $60\%$ fitness scaling for mine planning under uncertainty. Best, i.e. highest mean values are highlighted in {\textbf{bold face}}. Lower standard deviation values in comparison to approaches $(1)$ and $(2)$ are highlighted in \colorbox{gray!20} {color}. Lower standard deviation values in comparison to approaches $(1)$, $(2)$, $(3)$, and $(4)$ are highlighted in \colorbox{gray!70} {color}.
\label{tab:profit_results_I}
}
\vspace{3mm}
\begin{scriptsize}
\setlength{\tabcolsep}{9.3pt}  
\renewcommand{\arraystretch}{1.6} 
\begin{tabular}{rrrrrrrrrrrrrrrrr}
\toprule
\multicolumn{17}{c}{\textbf{NPV}}\\
\cmidrule(l{3pt}r{3pt}){2-17}

                      & \multicolumn{4}{c}{\textbf{Strategy (1)}}                                                                         & \multicolumn{4}{c}{\textbf{Uncertainty Discounted Strategy, 60\% (2)}}                                                         \\
\cmidrule(l{3pt}r{3pt}){2-5} \cmidrule(l{3pt}r{3pt}){6-9}
\cmidrule(l{3pt}r{3pt}){10-13} \cmidrule(l{3pt}r{3pt}){14-17}

\multicolumn{1}{c}{\textbf{p}} & \multicolumn{1}{c}{\textbf{max}} & \multicolumn{1}{c}{\textbf{min}} & \multicolumn{1}{c}{\textbf{mean $E[X]$}} & \multicolumn{1}{c}{\textbf{std}} & \multicolumn{1}{c}{\textbf{max}} & \multicolumn{1}{c}{\textbf{min}} & \multicolumn{1}{c}{\textbf{mean $E[X]$}} & \multicolumn{1}{c}{\textbf{std}} & \\ \hline
1 & 46914274.46 & 26955739.89 & 26955739.89& \cellcolor{gray!20}\textbf{6369614.01}&
49407751.35 & 28120950.59 &\textbf{38764350.97}& 6793520.03&
\\

2 & 134907416.40 & 113352199.10 & 124129807.70 & \cellcolor{gray!20}\textbf{6879183.13}  &
137600371.50&114610626.60&\textbf{126105499.10}&7337001.63 &
\\

3 & 119660703.90  & 118212619.70 & 118936661.80 & 462145.01 &
121121011.20&121019805.60&\textbf{121070408.40}&\cellcolor{gray!70}\textbf{32298.97} &
\\

4 & 127447435.70 & 101432494.90 & 114439965.30 & 8302469.84  &
128915265.60&104573557.70&\textbf{116744411.60}&\cellcolor{gray!20}\textbf{7768470.35}
 \\

5 & 124698872.90 & 71029797.81 & \textbf{97864335.33} & 17128075.77 &
97193249.16&45331265.59&71262257.38&\cellcolor{gray!20}\textbf{16551356.31}  &
 \\

6 & 135862400.80 & 54294009.92 & \textbf{95078205.36} & 26031929.52  &
106156327.20&26539595.13&66347961.17&\cellcolor{gray!20}\textbf{25409072.51}  &
\\

7 & 117950948.50 & 55430025.32 & 86690486.91& \cellcolor{gray!20}\textbf{19953075.55} &
128051455.10&59941197.48&\textbf{93996326.30}&21736869.00  &
 \\

8 & 118083180.30 & 74166373.14 & \textbf{96124776.74} & \cellcolor{gray!70}\textbf{14015713.90}  &
91543247.26&43557107.98&67550177.62&15314410.18  \\

9 & 72241870.40 & 50384657.60 & 61313264.00 & \cellcolor{gray!70}\textbf{6975562.68} &
99894321.05&69067936.93&\textbf{84481128.99}&9838005.26
 \\

10 & -& - & -& - &
26182969.26&18463413.38&\textbf{22323191.32}&\cellcolor{gray!70}\textbf{2463637.35} 
 \\

\bottomrule
\end{tabular}
\end{scriptsize}
\label{tab:resultsNPV}
\end{table*}

\begin{table*}[!t]

\caption{\label{tab:2} 
Maximum (\text{max}), minimum (\text{min}), mean (\text{mean}), and standard deviation (\text{std}) in terms of expected value for NPV $(\$)$ for uncertainty discounted Strategy with a $8\%$ discount rate, and a $90\%$ and a $99\%$ fitness scaling for mine planning under uncertainty. Best, i.e. highest mean values are highlighted in {\textbf{bold face}}. Lower standard deviation values in comparison to approaches $(1)$ and $(2)$ are highlighted in \colorbox{gray!20} {color}. Lower standard deviation values in comparison to approaches $(1)$, $(2)$, $(3)$, and $(4)$ are highlighted in \colorbox{gray!70} {color}.
\label{tab:profit_results_II}
}
\centering
\vspace{3mm}
\begin{scriptsize}
\setlength{\tabcolsep}{9.3pt} 
\renewcommand{\arraystretch}{1.6} 

\begin{tabular}{rrrrrrrrrrrrrrrrr}
\toprule
\multicolumn{17}{c}{\textbf{NPV}}\\
\cmidrule(l{3pt}r{3pt}){2-17}

                      & \multicolumn{4}{c}{\textbf{Uncertainty Discounted Strategy, 90\% (3)}}                                                                         & \multicolumn{4}{c}{\textbf{Uncertainty Discounted Strategy, 99\% (4)}}                                                                               \\
\cmidrule(l{3pt}r{3pt}){2-5} \cmidrule(l{3pt}r{3pt}){6-9}
\cmidrule(l{3pt}r{3pt}){10-13} \cmidrule(l{3pt}r{3pt}){14-17}

\multicolumn{1}{c}{\textbf{p}} & \multicolumn{1}{c}{\textbf{max}} & \multicolumn{1}{c}{\textbf{min}} & \multicolumn{1}{c}{\textbf{mean $E[X]$}} & \multicolumn{1}{c}{\textbf{std}} & \multicolumn{1}{c}{\textbf{max}} & \multicolumn{1}{c}{\textbf{min}} & \multicolumn{1}{c}{\textbf{mean $E[X]$}} & \multicolumn{1}{c}{\textbf{std}} & \\ \hline
1 & 33537895.59 &12940714.40 & \textbf{23239305.00} & 6573433.21&
31097623.29&
11619881.40&
21358752.35&
\cellcolor{gray!70}\textbf{6216172.70}\\
2 & 120460927.20 & 98215971.55 & \textbf{109338449.40} & 7099307.86 &
117825433.20&
96789471.91&
107307452.50&
\cellcolor{gray!70}\textbf{6713466.51}\\ 
3 & 103313578.10& 102610411.40 & 102961994.80& \cellcolor{gray!20}\textbf{224410.30} &
122609658.10&
119562311.40&
\textbf{121085984.70}&
972537.43\\ 
4 & 109792539.90 & 84582109.93& 97187324.93 & 8045716.36  &
110712954.60&
88431150.01&
\textbf{99572052.32}&
\cellcolor{gray!70}\textbf{7111067.92}\\ 
5 & 190483253.40 & 159750696.70 & 175116975.10 & 9808060.98 &
189967927.30&
161969280.10&
\textbf{175968603.70}&
\cellcolor{gray!70}\textbf{8935554.63}\\ 
6 & 139327276.00& 80563175.71 & 109945225.90 & 18754114.19 &
148629611.80&
92818133.77&
\textbf{120723872.80}&
\cellcolor{gray!70}\textbf{17811807.34}\\ 
7 & 159258982.10& 95456437.03 & 127357709.60 & 20362095.40 &
170716817.80&
110309278.30&
\textbf{140513048.10}&
\cellcolor{gray!70}\textbf{19278605.30}\\ 
8 & 121682529.10 & 66362446.19  & 94022487.66 & 17654982.34 &
134614408.40&
83144309.77&
\textbf{108879359.10}&
\cellcolor{gray!20}\textbf{16426289.24}\\ 
9 & 124184454.90 & 80267647.72  & \textbf{102226051.30} & \cellcolor{gray!20}\textbf{14015713.90} &
87676602.50&
42186928.24&
64931765.37&
14517682.42\\ 
10 & 72241870.40& 50384657.60 &61313264.00 & \cellcolor{gray!20}\textbf{6975562.68}  &
77661595.21&
54088014.20&
\textbf{65874804.71}&
7523328.50\\ 

\bottomrule
\end{tabular}
\end{scriptsize}
\label{tab:resultsNPV_Real_World_Data}
\end{table*}

\section{Applications to a test model}
\label{sec:test_model}
A test model was created to examine the effects of these changes on the fitness function. This model was introduced and illustrated in Section~\ref{sec:key} and Figure~\ref{fig:quality}. Although artificial, the aim of the test model is to show the new fitness function leading to schedules that open both pits simultaneously and attempt to blend material between the two pits at the expense of the ensemble-average NPV.

\subsection{Results using standard Evolution}
The test model was scheduled using standard Evolution Strategy and the average of three runs was taken. The mean value across the ensemble was used. The discount rate was set at $8\%$. As expected, stage-bench-domain combinations were mined in order of best early-access to value, leading to the entire East pit being mined followed by the entire West pit being mined. The tests produced $9$-year schedules with an average NPV of $\$36.935$ million and a total Standard Variance of $\$50.5$ million.

The resulting mining sequence was used to economically evaluate each ensemble member, producing a spread of $10$ solutions. For these tests we looked at the profit variance which showed large fluctuations in every period - except the first where only waste was mined. These fluctuations ranged from $\$21.58$ million to $\$23.68$ million. The standard deviation calculated across the ensembles is detailed in Table \ref{tab:resultsNPV} (1).

\subsection{Results using Evolution with uncertainty discounting}
The same model was now scheduled using the full ensemble via the new fitness function that incorporates the uncertainty discounting. The time discounting rate of $8\%$ per period was unchanged. $K_{\alpha}$ was chosen based on a given ${\alpha}$ of $60\%$, $90\%$ and $99\%$, by setting $F_{\alpha}$ to its corresponding value of $0.25$, $1.28$ and $2.33$ respectively. The results of these tests are detailed in Table \ref{tab:resultsNPV} (2), and Table \ref{tab:profit_results_II} (3) \& (4).

The change in schedules produced was clear. They ran for one period longer and the average NPV was lower at $\$23.139$ million. The total Standard Variance was also reduced to $\$46.6$ million. These schedules differed from the previous tests by uncovering ore in both East and West pits early and then mining them in an interleaved fashion where possible during the lifetime of the mine.
The resulting mining sequence was again used to economically evaluate each ensemble member, producing a spread of $10$ solutions. The fluctuations in profit variance in some periods were reduced in comparison to standard Evolution. The biggest reductions were seen in periods $4$, $6$, $7$ \& $8$. As an example, period $6$ saw an average profit of $\$40.3$ million, and a profit variance of $\$12.58$ million.

As is expected with regular Strategy, the results in Table~\ref{tab:resultsNPV} (1) show a small ensemble NPV standard deviation in the early periods, with large deviations in the later periods. The effect of uncertainty discounting is small in the Uncertainty Discounting Strategy, $60\%$ (2) and Uncertainty Discounting Strategy, $90\%$ (3) tests. When looking at the Uncertainty Discounting Strategy, $99\%$ (4) tests - where $F_{\alpha}$ corresponded to a value of $2.33$ - there were reductions in ensemble NPV standard deviation across several of the early and mid-schedule periods.

\subsection{Discussion of test model results}
Our tests showed that despite the additional period in the discounting approach, the total Standard Variance measure shows a reduction of approximately $\$4$ million. This implies that during the evolutionary process, schedules have been selected that reduce uncertainty risk. As expected, these schedules come at the expense of reduced ensemble-average NPV. We observed the highest maximum NPV value at the period $t = 9$ using uncertainty discounted Strategy and $90\%$ fitness scaling. In contrast, we observe the smallest minimum NPV value at the period $t = 1$ using uncertainty discounted Strategy and $99\%$ fitness scaling.
Our tests also showed that schedules focusing only on ensemble-average NPV as a measure of success can be misleading. When uncertainty information is used to evaluate the fluctuations possible in such schedules, per-period profits can differ from the expected value by 10s of millions of dollars. In a real-world scenario, these kinds of outcomes result in investor expectation mismatches and can add significant overall financial risk to an operation.

Finally, our suite of tests proved that our approach is reducing downside risk by reducing the variance between each of the ensemble members - a sort of levelling effect. Because of the small size of the model - and thus the small number of ore blocks accessible at any one time - the effect of the Uncertainty Discounting approach at the mid-to-low end of $F_{\alpha}$ did not show much improvement. This encouraged further tests on a larger, more realistic model.

\begin{figure*}[t]
\centering
\includegraphics[width=1\textwidth]{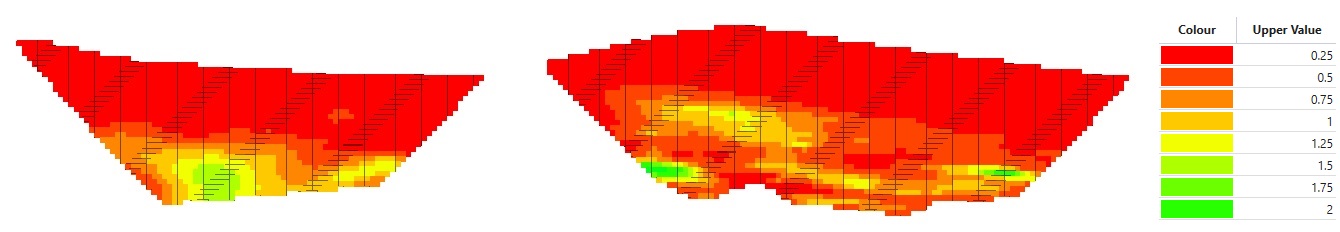}
\caption{(Complex model) A cross section through the Y plane of the , showing the distribution of copper grade - values in percent (\%) of block total.}
\label{fig:quality2}
\end{figure*}

\begin{table*}[!t]
\begin{small}
\caption{\label{tab:3} Maximum (\text{max}), minimum (\text{min}), mean (\text{mean}), and standard deviation (\text{std}) in terms of expected value for NPV (\$) for Strategy with a $8\%$ discount rate and uncertainty discounted Strategy with a $8\%$ discount rate and a $60\%$ fitness scaling for mine planning under uncertainty for real world model. Best, i.e. highest mean values are highlighted in {\textbf{bold face}}. Lower standard deviation values in comparison to approaches $(1)$ and $(2)$ are highlighted in \colorbox{gray!20} {color}. Lower standard deviation values in comparison to approaches $(1)$, $(2)$, $(3)$, and $(4)$ are highlighted in \colorbox{gray!70} {color}.
\label{tab:profit_results_real_world_V}
}
\centering
\vspace{3mm}
\begin{scriptsize}
\setlength{\tabcolsep}{9.3pt} 
\renewcommand{\arraystretch}{1.6} 

\begin{tabular}{rrrrrrrrrrrrrrrrr}
\toprule
\multicolumn{17}{c}{\textbf{NPV}}\\
\cmidrule(l{3pt}r{3pt}){2-17}

                      & \multicolumn{4}{c}{\textbf{Strategy (1)}}                                                                         & \multicolumn{4}{c}{\textbf{Uncertainty Discounted Strategy, 60\%(2)}}                                             \\
\cmidrule(l{3pt}r{3pt}){2-5} \cmidrule(l{3pt}r{3pt}){6-9}
\cmidrule(l{3pt}r{3pt}){10-13} \cmidrule(l{3pt}r{3pt}){14-17}

\multicolumn{1}{c}{\textbf{p}} & \multicolumn{1}{c}{\textbf{max}} & \multicolumn{1}{c}{\textbf{min}} & \multicolumn{1}{c}{\textbf{mean $E[X]$}} & \multicolumn{1}{c}{\textbf{std}} & \multicolumn{1}{c}{\textbf{max}} & \multicolumn{1}{c}{\textbf{min}} & \multicolumn{1}{c}{\textbf{mean $E[X]$}} & \multicolumn{1}{c}{\textbf{std}} & \\ \hline
1 & 401755011.10&
83034701.02&
\textbf{258203267.30}&
81022207.56&
280604267.20&
-16758146.98&
134732665.60&
\cellcolor{gray!20}\textbf{73094345.17}\\ 
2 & 591500398.50&
255627496.30&
\textbf{443844479.80}&
86567345.09 &
426888055.00&
107921951.00&
275299500.20&
\cellcolor{gray!20}\textbf{77541435.56}\\ 
3 & 617648621.10&
284977602.20&
\textbf{468796029.70}&
86546728.47&
441405528.30&
106040865.10&
300262779.50&
\cellcolor{gray!20}\textbf{78851752.22}\\ 
4 & 691244275.90&
325355764.10&
\textbf{542476305.40}&
90448845.01&
521086196.10&
191385033.10&
382142755.30&
\cellcolor{gray!20}\textbf{80713567.01}\\ 
5 & 860583098.10&
491396468.90&
\textbf{713852699.70}&
88967371.65 &
677585422.30&
348286673.60&
541446204.40&
\cellcolor{gray!70}\textbf{77156323.62}\\ 
6 & 1129563920.00&
731173535.70&
\textbf{970927647.20}&
95759149.63 &
940482789.20&
585401895.50&
793787156.10&
\cellcolor{gray!70}\textbf{83307669.13}\\ 
7 & 1061910707.00&
691765164.40&
894686741.30&
\cellcolor{gray!20}\textbf{84140534.03}  &
1069495618.00&
694318126.40&
\textbf{919092261.40}&
93257454.37\\ 
8 & 656367774.90&
286156724.30&
517323057.40&
\cellcolor{gray!70}\textbf{73306656.11}&  
1361714821.00&
956249642.70&
\textbf{1199171778.00}&
100593574.00\\ 
9 & 919477196.90&
519649262.20&
769308902.00&
\cellcolor{gray!70}\textbf{79171188.60} &
1582882707.00&
1148438216.00&
\textbf{1399130344.00}&
103306290.60\\ 
10 & 1189688495.00&
759637664.70&
1027286990.00&
\cellcolor{gray!20}\textbf{85038630.49}&
1370989815.00&
919053198.20&
\textbf{1180649244.00}&
97216529.47\\ 
11&1159650466.00&
796735973.10&
989916178.90&
\cellcolor{gray!70}\textbf{81481387.60} &
1290688622.00&
855800945.60&
\textbf{1136335595.00}&
84358036.72\\ 
12&1026802858.00&
698294568.80&
884610656.50&
\cellcolor{gray!70}\textbf{72628478.06} &
1469042908.00&
1018370350.00&
\textbf{1302942247.00}&
90148103.14\\ 
13&1292297919.00&
940774583.10&
1138250851.00&
\cellcolor{gray!70}\textbf{78025864.71}&
1348752455.00&
949083164.20&
\textbf{1165148673.00}&
86421590.04\\ 
14&1395317157.00&
999329973.00&
\textbf{1223200323.00}&
\cellcolor{gray!70}\textbf{84459868.23}&
1398475590.00&
992691540.60&
1222556590.00&
86188712.11\\ 
15&1403536310.00&
1012694410.00&
\textbf{1223600010.00}&
\cellcolor{gray!70}\textbf{86332882.35} &
1399724610.00&
998682422.00&
1216169330.00&
88353113.77\\ 
16&1329910243.00&
936086873.70&
\textbf{1134362236.00}&
\cellcolor{gray!70}\textbf{87292364.11}&
1312015311.00&
908726604.50&
1113537182.00&
89548598.62\\ 
17& 
1129513002.00&
729933858.60&
\textbf{935476426.50}&
\cellcolor{gray!70}\textbf{86829102.74}&
1096947304.00&
681127266.00&
900995707.00&
90204033.23\\ 
18&832571043.20&
395979676.90&
\textbf{599341018.80}&
86132506.71&
779841779.80&
422036469.00&
551484740.90&
\cellcolor{gray!20}\textbf{78414444.83}\\
19&530977086.50&
183093780.30&
\textbf{358951698.20}&
\cellcolor{gray!70}\textbf{68934813.07}&
502963318.00&
123113536.30&
325721191.90&
76742563.31\\
20&216795703.40*&
17652997.39&
\textbf{81019888.32}&
\cellcolor{gray!20}\textbf{91442845.67}&
153754939.30*&
175681.42&
76965310.38&
108596934.70\\

\bottomrule
\end{tabular}
\end{scriptsize}
\label{tab:resultsNPV_Real_World_Data_8}
\end{small}
\end{table*}

\begin{table*}[!t]
\small
\caption{\label{tab:4} Maximum (\text{max}), minimum (\text{min}), mean (\text{mean}), and standard deviation (\text{std}) in terms of expected value for NPV (\$) for uncertainty discounted Strategy with a $8\%$ discount rate and $90\%$ and $99\%$ fitness scaling for mine planning under uncertainty for real world model. Best, i.e. highest mean values are highlighted in {\textbf{bold face}}. Lower standard deviation values in comparison to approaches $(3)$ and $(4)$ are highlighted in \colorbox{gray!20} {color}. Lower standard deviation values in comparison to approaches $(1)$, ($2)$, $(3)$, and $(4)$ are highlighted in \colorbox{gray!70} {color}.
\label{tab:profit_results_real_world_VI}
}
\centering
\vspace{3mm}
\begin{scriptsize}
\setlength{\tabcolsep}{9.3pt} 
\renewcommand{\arraystretch}{1.6} 

\begin{tabular}{rrrrrrrrrrrrrrrrr}
\toprule
\multicolumn{17}{c}{\textbf{NPV}}\\
\cmidrule(l{3pt}r{3pt}){2-17}

                      & \multicolumn{4}{c}{\textbf{Uncertainty Discounted Strategy, 90\%( (3)}}                                                                         & \multicolumn{4}{c}{\textbf{Uncertainty Discounted Strategy, 99\%(4)}}                                                                               \\
\cmidrule(l{3pt}r{3pt}){2-5} \cmidrule(l{3pt}r{3pt}){6-9}
\cmidrule(l{3pt}r{3pt}){10-13} \cmidrule(l{3pt}r{3pt}){14-17}

\multicolumn{1}{c}{\textbf{p}} & \multicolumn{1}{c}{\textbf{max}} & \multicolumn{1}{c}{\textbf{min}} & \multicolumn{1}{c}{\textbf{mean $E[X]$}} & \multicolumn{1}{c}{\textbf{std}} & \multicolumn{1}{c}{\textbf{max}} & \multicolumn{1}{c}{\textbf{min}} & \multicolumn{1}{c}{\textbf{mean $E[X]$}} & \multicolumn{1}{c}{\textbf{std}} & \\ \hline
1 & 265575303.30&
-33787409.31&
\textbf{125942199.90}&
73412362.00  &
184510599.60&
-85823269.28&
44890779.35&
\cellcolor{gray!70}\textbf{67855579.40}\\ 
2 & 448518102.20&
133365131.40&
\textbf{304662954.60}&
78221250.00 &
409871447.60&
117910869.20&
259082041.70&
\cellcolor{gray!70}\textbf{73284025.75}\\ 
3 & 466358727.50&
121815015.20&
\textbf{326724340.70}&
80995369.53 &
382097074.20&
70891950.30&
241554264.10&
\cellcolor{gray!70}\textbf{74820467.90}\\ 
4 & 520628810.50&
176354127.10&
\textbf{377134332.20}&
81289502.96&
424892996.40&
116584340.10&
283161657.60&
\cellcolor{gray!70}\textbf{77841937.83}\\ 
5 & 771356731.00&
400919572.00&
\textbf{617337071.50}&
87723735.61 &
669237656.80&
336324236.10&
516210056.90&
\cellcolor{gray!20}\textbf{84068470.17}\\ 
6 & 950142032.30&
580352808.50&
\textbf{797958887.20}&
\cellcolor{gray!20}\textbf{85788280.92}  &
889971998.20&
537702581.00&
732439498.10&
90217781.63\\ 
7 & 1234916930.00&
835118306.60&
\textbf{1070432319.00}&
92624674.14  &
963186909.50&
621186581.20&
814244293.70&
\cellcolor{gray!70}\textbf{89392762.07}\\ 
8 & 1484261300.00&
1063005099.00&
\textbf{1310534946.00}&
98150019.25&  
1250841862.00&
881481507.70&
1089983837.00&
\cellcolor{gray!20}\textbf{96544183.03}\\ 
9 & 1329494343.00&
897359689.90&
1137543526.00&
\cellcolor{gray!20}\textbf{93004285.33} &
1557465140.00&
1159135881.00&
\textbf{1384007502.00}&
104248969.10\\ 
10 & 958814953.70&
551017324.10&
809269696.30&
\cellcolor{gray!70}\textbf{78658388.46}&
1655449321.00&
1191387663.00&
\textbf{1474195010.00}&
108104064.90\\ 
11&1246120150.00&
805698710.00&
1084395373.00&
\cellcolor{gray!20}\textbf{85080961.48} &
1500947259.00&
1044645974.00&
\textbf{1320031746.00}&
104095875.20\\ 
12& 1444404060.00&
991296056.50&
1268439124.00&
\cellcolor{gray!20}\textbf{89404737.69}&
1811632476.00&
1322650878.00&
\textbf{1616836343.00}&
112103052.70\\ 
13&1231344158.00&
790716720.10&
1042318979.00&
\cellcolor{gray!20}\textbf{97991706.01}&
1719165420.00&
1231154679.00&
\textbf{1507653547.00}&
111169726.50\\
14& 1374762701.00&
972125463.40&
1199663538.00&
\cellcolor{gray!20}\textbf{86024422.84}&
1379262124.00&
976241700.70&
\textbf{1202992352.00}&
86708950.54\\ 
15& 1381998329.00&
983139205.80&
\textbf{1198928539.00}&
\cellcolor{gray!20}\textbf{88856388.14}&
1379878945.00&
980402884.50&
1196227137.00&
89619446.74\\ 
16&1287697853.00&
889564153.00&
1090413972.00&
\cellcolor{gray!20}\textbf{89273175.87}&
1292386716.00&
888666761.00&
\textbf{1090664520.00}&
91361316.02\\ 
17& 1048357582.00&
644284680.90&
857201873.90&
\cellcolor{gray!20}\textbf{91238472.09}&
1068234767.00&
648933232.90&
\textbf{868594302.90}&
92976913.99\\ 
18&706940089.30&
411675864.90&
\textbf{540886203.70}&
\cellcolor{gray!70}\textbf{61088549.55}&
712590420.10&
387410210.80&
503886289.30&
68319350.93\\
19& 474381541.50*&
15178705.60&
\textbf{277895680.30}&
90184092.36&
440632047.90*&
11646856.23&
253594083.50&
\cellcolor{gray!20}\textbf{87938948.03}\\

\bottomrule
\end{tabular}
\end{scriptsize}
\label{tab:resultsNPV_Real_World_Data_8-2}
\end{table*}

\section{Application to a more complex model}
\label{sec:complex_model}
A large complex model consisting of over $200\,000$ blocks was now created to represent a real-world disseminated copper deposit. The model is shown in Figure~\ref{fig:quality2}. This benchmark dataset is available for study\footnote{Downloadable .zip file at \url{https://tiny.cc/mcm_zip}}.
Some characteristics remain in common with the test model in that there are two main pits required to mine the ore. The ore uncovered by the East pit comprises a large volume of medium grade material, averaging $0.51\%$ copper, with some material relatively close to the surface. The West pit holds deeper ore as smaller lenses of higher grade material. Because of its depth and the amount of waste material that needs to be moved, its grade averages $0.43\%$. When looking exclusively at blocks that contain at least $0.25\%$ copper - a common minimum grade chosen to delineate ore from waste - the East deposit averages $0.61\%$, while the West pit averages $0.64\%$.

The variability in grade distribution is captured in $50$ ensemble members. The method used to generate the uncertainty is artificial but approximates real world drilling uncertainties by sampling a `ground truth' model using increasing sampling distances and spatial uncertainty with depth. The spatial uncertainties are random but correlated at approximately $5$ blocks ($50$m scale). This results in domain boundary uncertainty - the boundary between the deposit and the surrounding barren rock as well as over estimation and underestimation of ore grades on approximately a +/-$50$m scale from member to member. The resulting differences between members is non-linear but spatially correlated. Typically grade values vary by +/-$20$\% across the ensemble for some blocks but can be substantially more in others - for example those near a domain boundary.

Unfortunately a real world deposit with $50$ realisations of the geology was not available for publication but they are produced by the geology teams in some operations. While the source of the geological uncertainty is not the primary interest here, the method used attempts to provide a substitute equal in complexity to a real world deposit. The scenario, even in the absence of variability information, provides the mining engineer with multiple options on how to schedule extraction - ranging from sequential phasing through to having multiple pit and pit faces open simultaneously. 

\subsection{Results using standard Evolution}
Three tests were run using the same setup information and the results averaged. By-and-large, the deposit was sequentially mined. The West pit was mined first as it contained higher grade material in a smaller area. While portions of the East pit were selectively mined to remove waste - known as pre-stripping - the primary focus was on the West pit, before moving onto the East pit. The asterisk indicates that in this period not all data were available.
The tests produced schedules with an average NPV of $\$394.87$ million over $20$ years (periods), and an average total Standard Variance of $\$390$ million. The results of these tests are detailed in Table \ref{tab:resultsNPV_Real_World_Data_8} (1).
Each ensemble member was evaluated using the resulting mining sequence, showing the spread of possible solutions. For these tests we looked at the profit variance which showed large fluctuations in periods $6$, $7$, $11$, $19$ \& $20$ - at some points showing differences of $\$324.5$ million.

\subsection{Results using Evolution with uncertainty discounting}
Three tests were run using the same basic setup information as the previous tests. Again, the primary objective was for Strategy's stage-bench-domain ranking to provide an order based on value, however this time that value was discounted using our new uncertainty discounting approach - the same as our tests for the test model. The intended result is for benches from each pit to be mined in the same period. The larger area of this complex model in comparison to the test model provides a larger number of potential areas to mine from, allowing for more options for the algorithm to reduce uncertainty.
The time discounting rate of $8\%$ per period was unchanged. $K_{\alpha}$ was chosen based on a given ${\alpha}$ of $60\%$, $90\%$ and $99\%$, by setting $F_{\alpha}$ to its corresponding value of $0.25$, $1.28$ and $2.33$ respectively. The results of these tests are detailed in Table \ref{tab:resultsNPV_Real_World_Data_8} (2), and Table \ref{tab:resultsNPV_Real_World_Data_8} (3) \& (4).

The tests produced schedules with an average NPV of $\$264.94$ million, completing in between $19$ and $20$ years (periods), with an average Standard Variance of $\$349.24$ million. The decrease in total NPV is directly related to attempting to simultaneously mine benches from both pits, resulting in mining ore later in the schedule. This again shows that the optimisation engine is attempting to mix benches from alternate pits to try and blend away the uncertainty.

Fluctuations in per-period profit have been spread further into the schedule, with the largest fluctuations seen in periods $6$, $9$, $10$, $12$, $17$, $18$ \& $19$. The largest variance of \$438 million was in the final period. This was common when using fitness scaling and discount rate together, as higher uncertainty is pushed further back into the schedule where the time-value of money is lower. Although highest NPV value $\$1.60$B is achieved later in period $t = 12$ using uncertainty discounted Strategy and $99\%$ fitness scaling, we observe that the highest maximum NPV values are achieved from period $t = 8$ until $t = 16$. Morever, we observe the smallest minimum NPV value at the period $t = 1$ using uncertainty discounted Strategy and $99\%$ fitness scaling.
As shown in the results tables, the traditional Strategy approach (1) shows larger ensemble NPV standard variations in the later periods. In comparison, the uncertainty discounted Strategy, $60\%$ (2) and uncertainty discounted Strategy, $99\%$ (4) show smaller ensemble NPV standard deviations early on in the schedule. This supports our theory that the discounted approach reduces variation between the ensemble members, making any possible realisation of a single ensemble member less of a risk.

Some of the solutions using fitness scaling were able to complete the schedule in one less period due to the way ore and waste were mined simultaneously. In the traditional Strategy approach, ore is always mined first (where available) and sent direct to the mill. Once the mill has reached its yearly capacity, all mining stops - including any mining of waste.

\subsection{Conclusion from tests}
Our tests have shown to reduce downside risk by (1) blending uncorrelated areas of the pit together and by (2) pushing uncertain areas of the pit to later in the mining schedule.
In both scenarios, mining of the ore is done later in the schedule and as a result the NPV of schedules is lower. Our tests support our theory that the optimisation changes result in schedules with reduced risk, while providing more certainty around cash flow.
These tests did not consider any changes provided through alternative staging - choosing instead to focus on maximising uncertain value. As explained by Reid et al.~\cite{DBLP:conf/gecco/Reid0RN21}, staging has a huge effect on NPV and how uncertainty can be realised, and is another factor in what has evidently become a large and complex multi-objective problem.
Finally, our use of the larger model showed incorporating uncertainty into the fitness calculations - even for small $F_{\alpha}$ - could help to reduce the spread of potential profit in earlier periods. This pushed uncertain areas of mining to later periods, where the time value of money is lower and the effect of economic variation would not be as hard. Additionally, pushing uncertainty back in time can make sure that the return on investment date is hit on time so that a mine site can remain operational.

\section{Industry applications}
With these results in mind, we firmly believe the discounting process and associated results can be applied as further justification for improved accuracy in supporting existing drilling programs for sites, in any part of it's life cycle. It also assists mine planning engineers by better understanding any financial implications a lower realised grade may yield, further reducing downside risk.

The results of our work have been successfully shown our changes can minimize downside risk, at the cost of not achieving maximum upside profit. Our work should be seen as a way to visualise and mitigate portions of a deposit that are highly variable in grade, which could result in significant loss if the realised value is lower than initially estimated.

\section{Conclusions and Future work}
We have shown that our new approach based on uncertainty discounting reduces uncertainty in early periods of the mine scheduling process.
Our work will be included as part of the Maptek Evolution 2023 release, expected in mid-April. As it has only recently been completed we will be continuing our investigations on real-world pits to determine the level of impact our solution can provide. This work will also help us look at trade-offs in terms of expected value and variance.

It is expected that this work will assist in further changes to the algorithm, and educate further adjustment to fitness scaling values. When implementing this approach in the Evolution Strategy engine we included the constant term $K_{\alpha}$ described above. 
The effect of this value could be further exploited by providing a spread of solutions with increasing $K_{\alpha}$.

\section*{Acknowledgement}
This work has been supported by the South Australian Government through the Research Consortium "Unlocking Complex Resources through Lean Processing" and by the Australian Research Council through grant FT200100536.



\begin{thebibliography}{10}
\providecommand{\url}[1]{#1}
\csname url@samestyle\endcsname
\providecommand{\newblock}{\relax}
\providecommand{\bibinfo}[2]{#2}
\providecommand{\BIBentrySTDinterwordspacing}{\spaceskip=0pt\relax}
\providecommand{\BIBentryALTinterwordstretchfactor}{4}
\providecommand{\BIBentryALTinterwordspacing}{\spaceskip=\fontdimen2\font plus
\BIBentryALTinterwordstretchfactor\fontdimen3\font minus
  \fontdimen4\font\relax}
\providecommand{\BIBforeignlanguage}[2]{{%
\expandafter\ifx\csname l@#1\endcsname\relax
\typeout{** WARNING: IEEEtran.bst: No hyphenation pattern has been}%
\typeout{** loaded for the language `#1'. Using the pattern for}%
\typeout{** the default language instead.}%
\else
\language=\csname l@#1\endcsname
\fi
#2}}
\providecommand{\BIBdecl}{\relax}
\BIBdecl

\bibitem{bienstock2010solving}
D.~Bienstock and M.~Zuckerberg, ``Solving {LP} relaxations of large-scale
  precedence constrained problems,'' in \emph{{IPCO}}, ser. Lecture Notes in
  Computer Science, vol. 6080.\hskip 1em plus 0.5em minus 0.4em\relax Springer,
  2010, pp. 1--14.

\bibitem{DBLP:journals/coap/MunozEGMQL18}
G.~Mu{\~{n}}oz, D.~G. Espinoza, M.~Goycoolea, E.~Moreno, M.~Queyranne, and
  O.~R. Letelier, ``A study of the {B}ienstock-{Z}uckerberg algorithm:
  applications in mining and resource constrained project scheduling,''
  \emph{Comput. Optim. Appl.}, vol.~69, no.~2, pp. 501--534, 2018.

\bibitem{DBLP:journals/ior/LetelierEGMM20}
O.~R. Letelier, D.~G. Espinoza, M.~Goycoolea, E.~Moreno, and G.~Mu{\~{n}}oz,
  ``Production scheduling for strategic open pit mine planning: {A}
  mixed-integer programming approach,'' \emph{Oper. Res.}, vol.~68, no.~5, pp.
  1425--1444, 2020.

\bibitem{DBLP:journals/asc/GoodfellowD16}
R.~C. Goodfellow and R.~G. Dimitrakopoulos, ``Global optimization of open pit
  mining complexes with uncertainty,'' \emph{Appl. Soft Comput.}, vol.~40, pp.
  292--304, 2016.

\bibitem{DBLP:journals/heuristics/MontielD17}
L.~Montiel and R.~G. Dimitrakopoulos, ``A heuristic approach for the stochastic
  optimization of mine production schedules,'' \emph{J. Heuristics}, vol.~23,
  no.~5, pp. 397--415, 2017.

\bibitem{DBLP:journals/cor/LamghariD20}
A.~Lamghari and R.~G. Dimitrakopoulos, ``Hyper-heuristic approaches for
  strategic mine planning under uncertainty,'' \emph{Comput. Oper. Res.}, vol.
  115, p. 104590, 2020.

\bibitem{maptek}
{Maptek Pty. Ltd}, ``Maptek evolution,''
  \url{https://www.maptek.com/products/evolution/index.html}, 2023.

\bibitem{minemax}
{MineMax Pty. Ltd}, ``Minemax scheduler,''
  \url{https://www.minemax.com/products/scheduler/}, 2023.

\bibitem{DBLP:journals/jour/Ibrahimov2014}
M.~Ibrahimov, A.~Mohais, S.~Schellenberg, and Z.~Michalewicz, ``Scheduling in
  iron ore open-pit mining,'' \emph{The International Journal of Advanced
  Manufacturing Technology}, vol.~72, no.~5, pp. 1021--1037, 2014.

\bibitem{DBLP:conf/cec/OsadaWBM13}
Y.~Osada, L.~While, L.~Barone, and Z.~Michalewicz, ``Multi-mine planning using
  a multi-objective evolutionary algorithm,'' in \emph{{IEEE} Congress on
  Evolutionary Computation}.\hskip 1em plus 0.5em minus 0.4em\relax {IEEE},
  2013, pp. 2902--2909.

\bibitem{chakraborty2012multi}
A.~Chakraborty and M.~Chakraborty, ``Multi criteria genetic algorithm for
  optimal blending of coal,'' \emph{{OPSEARCH}}, vol.~49, no.~4, pp. 386--399,
  2012.

\bibitem{DBLP:conf/gecco/SchellenbergLM16}
S.~Schellenberg, X.~Li, and Z.~Michalewicz, ``Benchmarks for the coal
  processing and blending problem,'' in \emph{{GECCO}}.\hskip 1em plus 0.5em
  minus 0.4em\relax {ACM}, 2016, pp. 1005--1012.

\bibitem{DBLP:conf/gecco/XieN022}
Y.~Xie, A.~Neumann, and F.~Neumann, ``An optimization strategy for the complex
  large-scale stockpile blending problem,'' in \emph{{GECCO} Companion}.\hskip
  1em plus 0.5em minus 0.4em\relax {ACM}, 2022, pp. 770--773.

\bibitem{xie2021heuristic}
Y.Xie, A.~Neumann, and F.~Neumann, ``Heuristic strategies for solving complex
  interacting stockpile blending problem with chance constraints,'' in
  \emph{{GECCO}}.\hskip 1em plus 0.5em minus 0.4em\relax {ACM}, 2021, pp.
  1079--1087.

\bibitem{DBLP:conf/cec/Xie0N21}
Y.~Xie, A.~Neumann, and F.~Neumann, ``Heuristic strategies for solving complex
  interacting large-scale stockpile blending problems,'' in \emph{{IEEE}
  Congress on Evolutionary Computation, {CEC}}.\hskip 1em plus 0.5em minus
  0.4em\relax {IEEE}, 2021, pp. 1288--1295.

\bibitem{myburgh2010evolutionary}
C.~Myburgh and K.~Deb, ``Evolutionary algorithms in large-scale open pit mine
  scheduling,'' in \emph{{GECCO}}.\hskip 1em plus 0.5em minus 0.4em\relax
  {ACM}, 2010, pp. 1155--1162.

\bibitem{elsayed2020evolutionary}
S.~Elsayed, R.~Sarker, D.~Essam, and C.~A.~C. Coello, ``Evolutionary approach
  for large-scale mine scheduling,'' \emph{Information Sciences}, vol. 523, pp.
  77--90, 2020.

\bibitem{ferland2007application}
J.~A. Ferland, J.~Amaya, and M.~S. Djuimo, ``Application of a particle swarm
  algorithm to the capacitated open pit mining problem,'' in \emph{Autonomous
  Robots and Agents}.\hskip 1em plus 0.5em minus 0.4em\relax Springer, 2007,
  pp. 127--133.

\bibitem{gu2021layout}
Q.~Gu, X.~Li, L.~Chen, and C.~Lu, ``Layout optimization of crushing station in
  open-pit mine based on two-stage fusion particle swarm algorithm,''
  \emph{Engineering Optimization}, vol.~53, no.~10, pp. 1671--1694, 2021.

\bibitem{cressie1990origins}
N.~Cressie, ``The origins of kriging,'' \emph{Mathematical Geology}, vol.~22,
  no.~3, pp. 239--252, 1990.

\bibitem{kleijnen2009kriging}
J.~P. Kleijnen, ``Kriging metamodeling in simulation: A review,''
  \emph{European Journal of Operational Research}, vol. 192, no.~3, pp.
  707--716, 2009.

\bibitem{hoshiya1995kriging}
M.~Hoshiya, ``Kriging and conditional simulation of gaussian field,''
  \emph{Journal of Engineering Mechanics}, vol. 121, no.~2, pp. 181--186, 1995.

\bibitem{leuangthong2003stepwise}
O.~Leuangthong and C.~V. Deutsch, ``Stepwise conditional transformation for
  simulation of multiple variables,'' \emph{Mathematical Geology}, vol.~35,
  no.~2, pp. 155--173, 2003.

\bibitem{DBLP:conf/gecco/Reid0RN21}
W.~Reid, A.~Neumann, S.~Ratcliffe, and F.~Neumann, ``Advanced mine optimisation
  under uncertainty using evolution,'' in \emph{{GECCO} Companion}.\hskip 1em
  plus 0.5em minus 0.4em\relax {ACM}, 2021, pp. 1605--1613.

\bibitem{lerchs-grossmann}
H.~Lerchs and I.~F. Grossmann, ``Optimum design of open-pit mines,''
  \emph{Canadian Mining and Metallurgical Bulletin, CIM Bulletin, Montreal,
  Canada}, pp. 47--54, 1965.

\bibitem{caccetta2007application}
L.~Caccetta, ``Application of optimisation techniques in open pit mining,'' in
  \emph{Handbook of operations research in natural resources}.\hskip 1em plus
  0.5em minus 0.4em\relax Springer, 2007, pp. 547--559.

\bibitem{whittle1991open}
J.~Whittle and L.~Rozman, ``Open pit design in the 90s,'' in \emph{Proceedings
  Mining Industry Optimization Conference}, 1991.

\bibitem{DBLP:conf/ppsn/NeumannXN22}
A.~Neumann, Y.~Xie, and F.~Neumann, ``Evolutionary algorithms for limiting the
  effect of uncertainty for the knapsack problem with stochastic profits,'' in
  \emph{{PPSN} {(1)}}, ser. Lecture Notes in Computer Science, vol.
  13398.\hskip 1em plus 0.5em minus 0.4em\relax Springer, 2022, pp. 294--307.

\end{thebibliography}
\end{document}